\documentclass[10pt,twocolumn,letterpaper]{article}

\usepackage[pagenumbers]{cvpr/cvpr}      

\def\paperID{16} 
\def\confName{CVPR}
\def\confYear{2024}

\usepackage[utf8]{inputenc} 
\usepackage[T1]{fontenc}    
\usepackage{xr-hyper}
\usepackage[dvipsnames]{xcolor}

\definecolor{cvprblue}{rgb}{0.21,0.49,0.74}
\usepackage[pagebackref,breaklinks,colorlinks,citecolor=cvprblue]{hyperref}

\usepackage{url}            
\usepackage{booktabs}       
\usepackage{amsfonts}       
\usepackage{amsmath}
\usepackage{nicefrac}       
\usepackage{microtype}      
\usepackage{xcolor}         
\usepackage{mathtools}

\def\fref#1{Fig.~\ref{#1}}
\def\tref#1{Table~\ref{#1}}
\def\sref#1{Section~\ref{#1}}

\def\eref#1{Equation~\ref{#1}}

\makeatletter

\newcommand*{\addFileDependency}[1]{
\typeout{(#1)}
\@addtofilelist{#1}
\IfFileExists{#1}{}{\typeout{No file #1.}}
}\makeatother

\newcommand*{\myexternaldocument}[1]{%
\externaldocument{#1}%
\addFileDependency{#1.tex}%
\addFileDependency{#1.aux}%
}

\myexternaldocument{supplemental}

\title{Layered Diffusion Model for One-Shot High Resolution Text-to-Image Synthesis}

\author{%
  Emaad Khwaja\thanks{UC Berkeley - UCSF Joint Bioengineering Graduate Program} \ \thanks{Computer Science Division, UC Berkeley} \ \thanks{These authors contributed equally to this work.} \ \thanks{Work completed during an internship at Google.} \\
  {\tt\small emaad@berkeley.edu} \\
\and
  Abdullah Rashwan \footnotemark[3] \ \thanks{Google} \\
  {\tt\small arashwan@google.com} \\
  \and
    Ting Chen\thanks{Google Deepmind} \\
  {\tt\small iamtingchen@google.com} \\
  \and
    Oliver Wang\footnotemark[5] \\
  {\tt\small olwang@google.com} \\
  \and
Suraj Kothawade\footnotemark[5]\\
{\tt\small skothawade@google.com} \\
\and
    Yeqing Li\footnotemark[5] \\
{\tt\small yeqing@google.com} \\
}

\begin{document}

\maketitle

\begin{abstract}

We present a one-shot text-to-image diffusion model that can generate high-resolution images from natural language descriptions. Our model employs a layered U-Net architecture that simultaneously synthesizes images at multiple resolution scales. We show that this method outperforms the baseline of synthesizing images only at the target resolution, while reducing the computational cost per step. We demonstrate that higher resolution synthesis can be achieved by layering convolutions at additional resolution scales, in contrast to other methods which require additional models for super-resolution synthesis.

\end{abstract}

\section{Introduction} \label{introduction}

High-resolution synthesis of images in pixel space poses a challenge for diffusion models. Previous methods have addressed this by either learning in low-dimensional latent space \cite{rombach_high-resolution_2022, xue_raphael_2023} or generating images via cascaded models focused on increasing resolutions \cite{ramesh_hierarchical_2022, saharia_photorealistic_2022, balaji_ediff-i_2023}. However, these methods introduce additional complexity to the diffusion framework. \citet{hoogeboom_simple_2023} proposed a simpler diffusion model that synthesizes high-resolution images by using shifted noise scheduling and model scaling, among other optimizations. 

In this work, we extend the simple diffusion model framework to perform hierarchical image synthesis at multiple resolutions. Our goal is to design a light-weight model that can capture spatial image features at specific resolution scales within a singular model, rather than training with respect to the highest resolution. We argue that this approach is preferable to cascaded models, as it allows the upscaling process to be guided by a multitude of learned features within the network, rather than from singular output image pixels for each successive super-resolution model.

Recently, \citet{gu_matryoshka_2023} demonstrated a similar idea, termed "Matroyshka Diffusion." They report similar improvements when compared to multiple baseline approaches. Our model differs primarily in the use of scaled noise between resolutions (\sref{noise_scaling}) and training optimizations employed for faster convergence (\sref{training_optimizations}).

\section{Architecture}

\begin{figure*}[t]
\centering
\includegraphics[width=1\textwidth]{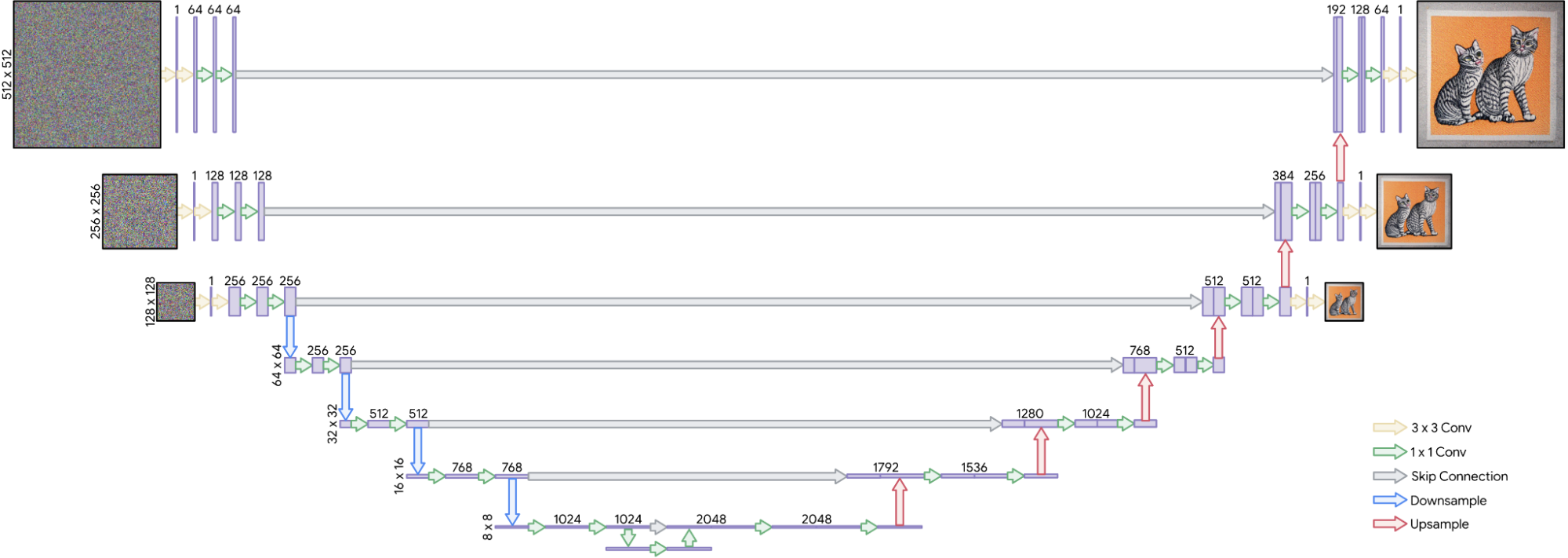}
\caption{Layered diffusion model architecture. We provide noisy inputs at 3 resolutions and generate images of corresponding size. Note that information is only shared between layers during upsampling via skip connection.}
\label{fig:architecture}
\end{figure*}

\subsection{Layered Model} \label{layered_model}
We introduce a new image generation architecture based on a U-Net \cite{ronneberger_u-net_2015} structure that produces images conditioned on text prompts at multiple resolution scales (\fref{fig:architecture}). Our architecture differs from a typical U-Net in a number of key ways:
\\
\begin{itemize}
    \item \textbf{Multiple Inputs}: At each resolution scale, we apply an input convolution to expand the input image to the hidden dimension size of the layer of the respective resolution scale.
    \\
    \item \textbf{Isolated Downsampling}: Within a standard U-Net architecture, downsampling convolutions are used to decrease tensor resolution. In our model, rather than downsample higher resolutions, the current tensor in the first part of the network, we utilize an input convolution of downscaled image using the technique described in \sref{noise_scaling}. The information within these tensors is only retained only as skip connections concatenated to the upsampling layers. However, at the lowest resolution ($128 \times 128$), the downsampling is allowed and the model operations proceed as normal (see \fref{fig:architecture}). We believe this encourages robust learning at every level rather than optimizing for high resolution features, which is important at the lowest resolution where image content is decided \cite{saharia_photorealistic_2022}. Additionally, eliminating high resolution convolutions decreases the total number of floating point operations (FLOPs) within the model.
    \\
    \item \textbf{Multiple Outputs}: We apply out convolutions at relevant resolution scales in the upsampling part of the network which are used to evaluate loss (mean squared error) during training. While higher resolution information is isolated during downsampling, it is concatenated with the output of the lower model layers via skip connection once upsampled. We scale loss with respect to resolution scale. During inference, only the output convolution of the highest resolution in necessary. 
    \\
\end{itemize}

Our model aims to learn spatial frequencies at different levels of the network. We adopt a hierarchical design philosophy for this model. We begin with a base model initially trained at a resolution of $128 \times 128$ with an embedding dimension of $256$. To include a higher resolution, we include largely parallel convolutional layers atop this structure with (e.g. $256 \times 256$ with hidden dimension of $128$). We show that our multi-scale model can generate higher quality images than a single resolution-scale model that directly generates images of the same resolution conditioned on text (where hidden dimension sizes are identical).

We compare models using Fréchet inception distance (FID) \cite{heusel_gans_2018} and Inception Score (IS) \cite{salimans_improved_2016} on the MS-COCO \cite{lin_microsoft_2015} validation set.

\subsection{Noise Scaling} \label{noise_scaling}
Rather than employing independently sampled Gaussian noise for each resolution scale, denoted as $\epsilon$, we explore a strategy which uses the sinc (also known as Whittaker-Shannon) interpolation formula whereby we randomly sample noise at the highest resolution, and downsample to the lower resolutions. Sinc interpolation acts as a lowpass filter to ensures the preservation of the noise signal across layers, allowing effective sharing of pixel information across different resolutions while maintaining the Gaussian nature of the noise \cite{whittaker_cardinal_1927}. We formulate our downsampling operation, $\mathcal{S}$, as:
\begin{equation*}
\hspace{-4.5cm} \mathcal{S}(\epsilon_0(\hat{m}, \hat{n}), T) = \sum_{m=0}^{M-1} \sum_{n=0}^{N-1}
\end{equation*}
\begin{equation} \label{eq:sinc_interpolation}
\hspace{1.25 cm}  \epsilon_0[m, n] (\text{sinc}\left(\frac{\hat{m} - mT}{T}\right) \cdot \text{sinc}\left(\frac{\hat{n} - nT}{T}\right))
\end{equation}

where $\epsilon_0$ represents noise randomly sampled at the highest resolution via $\epsilon \mathtt{\sim} \mathcal{N}(0, I)$, $M$ and $N$ are the dimensions of the target downscaled resolution, $\epsilon_0 [m,n]$ are the pixel values of the high resolution noise at location $(m,n)$, $(\hat{m}, \hat{n})$ denotes the corresponding pixel positions in the scaled noise, and $T$ is the downsampling scale. This formula represents a 2D version of the sinc interpolation formula applied to each noise term. 

We observed superior image synthesis when using sinc interpolation (Fig. A.2), with markedly lower FID and higher IS (Table A.1). We believe this approach is beneficial when compared to randomly sampled noise because pixel-level spatial frequencies for input noise are preserved at each resolution. While we use bilinear scaling for downsampling images (\eref{layered_objective_function}), applying the same operation to $\epsilon_0$ is problematic because bilinear interpolation nonlinearly modifies the signal profile and does not ensure that the output noise maintains a Gaussian distribution \cite{banelli_non-linear_2013}.
 \begin{table*}[!t]
  \centering
  \begin{tabular}{lclrr}
    \toprule
    \multicolumn{1}{c}{Model} & \multicolumn{1}{c}{Target Resolution} & \multicolumn{1}{c}{FLOPs} & \multicolumn{1}{c}{FID} & \multicolumn{1}{c}{IS} \\
    \midrule
    Single Resolution & $128 \times 128$ & $1.24 \times 10^{12}$ & $11.53$ & $24.48 \pm .41$ \\
    \midrule
    Single Resolution & $256 \times 256$ & $2.20 \times 10^{12}$ & $14.61$ & $28.61 \pm .44$ \\
    Layered & $256 \times 256$ & $2.04 \times 10^{12}$ & \textbf{13.38} & \textbf{29.62 $\pm$ .57} \\
    \midrule
    Single Resolution & $512 \times 512$ & $3.24 \times 10^{12}$ & $53.89$ & $23.21 \pm .29$ \\
    Layered & $512 \times 512$ & $2.79\times 10^{12}$ & \textbf{40.05} & \textbf{28.74 $\pm$ .47} \\
    \bottomrule
  \end{tabular}
\caption{ ``Target Resolution`` denotes the highest resolution of image used within the model. Models denoted as ``Single Resolution`` utilize a single resolution image at that scale in a typical U-Net architecture. ``Layered`` models utilize our multi-resolution model. Possible image inputs are $128 \times 128$, $256 \times 256$, and $512 \times 512$. Layered models are provided with multiple input images. For example, if the target resolution is $256 \times 256$, the layered model is provided with $128 \times 128$ and $256 \times 256$ input images.}
\label{table:simple_diffusion_comp}
\end{table*} 
\subsection{Cosine Schedule Shifting} \label{cosine_schedule_shifting}

\citet{hoogeboom_simple_2023} demonstrated that shifting the cosine schedules used to noise images helped overcome the obstacle of global image structure being defined for a long duration during image synthesis (see Fig. A.3). We similarly implement shifted cosine schedules for the higher resolution layers of the model (Fig. A.4). We first found the optimal delay for a layered model with highest resolution synthesis at $256 \times 256$ to be $log(\frac{1}{8})$. We then maintain this offset at the $256 \times 256$ resolution and swept to find the optimal delay for a model trained to synthesize at $512 \times 512$, which was $log (\frac{1}{32})$ (Fig. A.5). By doing so, our model is encouraged to synthesize fundamental image features earlier and incorporate textured information at higher levels later on in the diffusion process.

\subsection{Mathematical Formulation}

A text-to-image diffusion model $\hat{x}_{\theta}$ is trained on a denoising objective of given by:

\begin{equation} \label{objective_function}
\mathbb{E}_{x,c,\epsilon,t} \left[ w_t \left\| \hat{x}_\theta(\alpha_t x + \sigma_t \epsilon, c) - x \right\|^2_2 \right]
\end{equation}

where $(x, c)$ are data-conditioning pairs, $t \mathtt{\sim} \mathcal{U}([0,1])$,  and $\alpha_t$, $\sigma_t$, and $w_t$ are functions of $t$ that influence sample quality. 

To extend this objective function to multiple resolutions as discussed in \sref{layered_model}, we introduce a summation over different resolutions as:

\begin{equation*} 
\hspace*{-3cm} \mathbb{E}_{x_0,c,\epsilon_0,t} \Bigg[ w_t \sum_{i=0}^{K} \lVert  \hat{x}_{\theta,i}(\alpha_{t,i} \mathcal{D}(x_0, i) +
\end{equation*}
\begin{equation} \label{layered_objective_function}
\hspace*{3.5cm} \sigma_{t,i} \mathcal{S}(\epsilon_0, i), c) - \mathcal{D}(x_0, i) 
\rVert^2_2 \Bigg]
\end{equation}

where $K$ is the number of resolutions, $\hat{x}_{\theta, i}$ is the predicted image at resolution $i$, $\alpha_{t,i}$ and $\sigma_{t,i}$ are obtained via shifted cosine schedules described in \sref{cosine_schedule_shifting}, $\mathcal{D}(x_0, i)$ represents the bilinear downsampling operation applied directly to the highest resolution image $x_0$, and $\mathcal{S}(\epsilon_0, i)$ represents the sinc interpolation from \eref{eq:sinc_interpolation}.

\section{Training Optimizations} \label{training_optimizations}
\subsection{Strategic Cropping}
\citet{saharia_photorealistic_2022} demonstrated accelerated training convergence with respect to TPU time by training the $1024 \times 1024$ super-resolution model on $64 \times 64 \rightarrow 256 \times 256$ crops. We similarly found that we could utilize cropping to increase training speed by cropping all input images to $128 \times 128$. However, to accommodate our layered model architecture, a crop must also be applied within the a tensor immediately before upscaling (Fig. A.7). 
This was done with the following procedure:

\begin{enumerate}
\item Randomly select a $64 \times 64$ square within the base $128 \times 128$ image.
\item Identify the corresponding $128 \times 128$ positions within the $256 \times 256$ image via simple multiplication.
\item Crop $256 \times 256$ image to $128 \times 128$.
\item Repeat for higher resolutions by utilizing the previous resolution.
\item Apply the corresponding $64 \times 64$ crops in the upsample stack of the U-Net prior to the upsampling convolution to bring the output image to $128 \times 128$.
\end{enumerate}

\subsection{Model Stacking} \label{model_stacking}

A limitation of this implementation is that it requires images of at least the largest resolution of the model, which fundamentally limits the training set size based on depending on the highest resolution present. For example, a model trained on images of $[512 \times 512, 256 \times 256, 128 \times 128]$ would require a minimum base image resolution of $512 \times 512$

To address this issue, we leveraged the hierarchical nature of our model by iterative rounds of training and checkpointing. Each resolution scale is constructed upon its predecessor, allowing us to initialize a higher resolution model with weights from a model trained with a smaller maximum resolution.

We visualize our results in Table A.2. At $256 \times 256$, we compare a randomly initialized model (a) and a model where the lower weights are initialized with a $128 \times 128$ model trained to 500k steps. Similarly at $512 \times 512$, we compare a randomly initialized model with ones initialized with the (a) and (b) $256 \times 256$ models trained to 500k steps.

 Interestingly, we observed an increase in FID when using randomly initialized models. Despite this, our images demonstrated enhanced text-image alignment when the models were pre-loaded (Fig. A.7). More exploration is required to balance the image fidelity-prompt alignment trade off.
 
\section{Results}

In \tref{table:simple_diffusion_comp}, we evaluate our layered diffusion model against baseline models trained at singular target resolution from the perspective of image generation quality (FID and IS) as well as total floating point operations (FLOPs). We observe that our multi-scale synthesis outperforms single-scale models when the maximum output resolution was both $256 \times 256$ and $512 \times 512$ Note that the $256 \times 256$ comparison is evaluated on a model trained with on two input images [$128 \times 128$, $256 \times 256$], and not the individual outputs of the $512 \times 512$ model.) 

At $512 \times 512$ resolution, our model reduces the FID score by 25\% and increases the IS score by 24\%, while also decreasing the FLOPs by 14\%. This indicates that our model can generate more realistic and diverse images with less computation. Our multi-scale model also improves the FID and IS scores at $256 \times 256$ resolution by 8\% and 4\%, respectively, compared to the single-scale model. We observe image features in accordance with these findings (Fig. A.1), where higher resolution features can be seen in outputs from the layered model.

\section{Conclusion}

In this study, we propose a novel one-shot text-to-image diffusion model that can generate high-resolution images from natural language descriptions in a single forward pass. Our model employs a layered U-Net architecture that predicts images at multiple resolution scales simultaneously, using a sinc interpolation formula to scale noise noise and a shifted cosine schedule for the diffusion steps. We show that this method outperforms the baseline of synthesizing images only at the target resolution, while reducing the computational cost per step. Moreover, our model does not need any additional super-resolution models to enhance the image quality. We also explore various techniques to improve the training efficiency and the image quality, such as cropping and stacking. We believe that our model is an interesting path to explore further which streamlines text-to-image synthesis to a one-shot method.

{
    \small
    \bibliographystyle{cvpr/ieeenat_fullname}
    \bibliography{references}
}

\end{document}


\appendix

\begin{center}{\large \scshape \soptitle}\end{center}

\section{Introduction} \label{introduction_supp}

We compare our layered model to single resolution diffusion models in \fref{table:noise_scaling}. All models in this study are trained for 500k steps when evaluated.

\begin{figure}[H]
\centering
\includegraphics[width=1\textwidth]{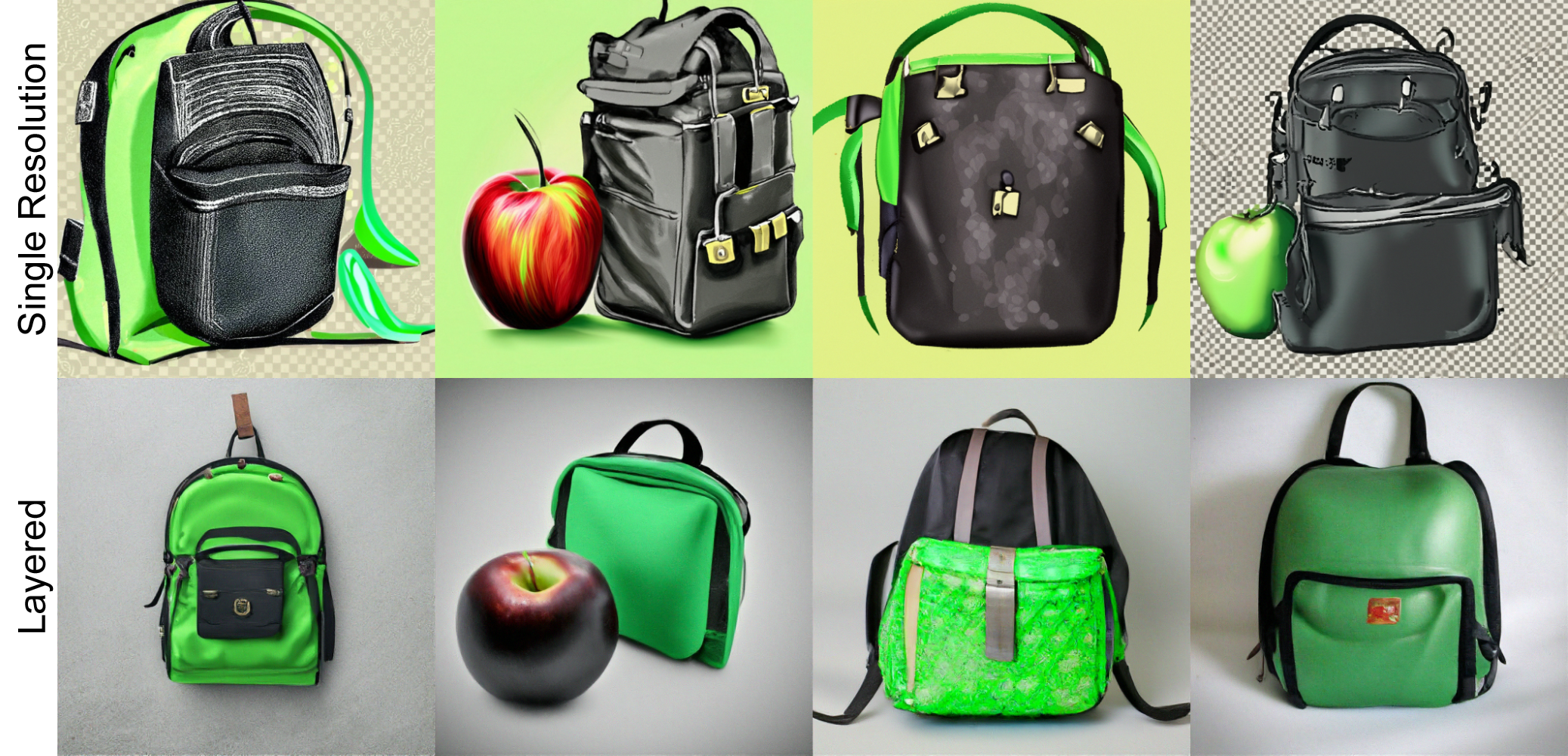}
\cprotect \caption{$512 \times 512$ outputs from prompt \textit{A black apple and a green backpack.} Finer detail textures can be seen in the layered model when compared with the single resolution model.}
\label{fig:single_versus_layered}
\end{figure}

\section{Architecture} \label{Architecture_supp}

\subsection{Noise Scaling} \label{noise_scaling_supp}

\begin{figure}[H]
\centering
\includegraphics[width=1\textwidth]{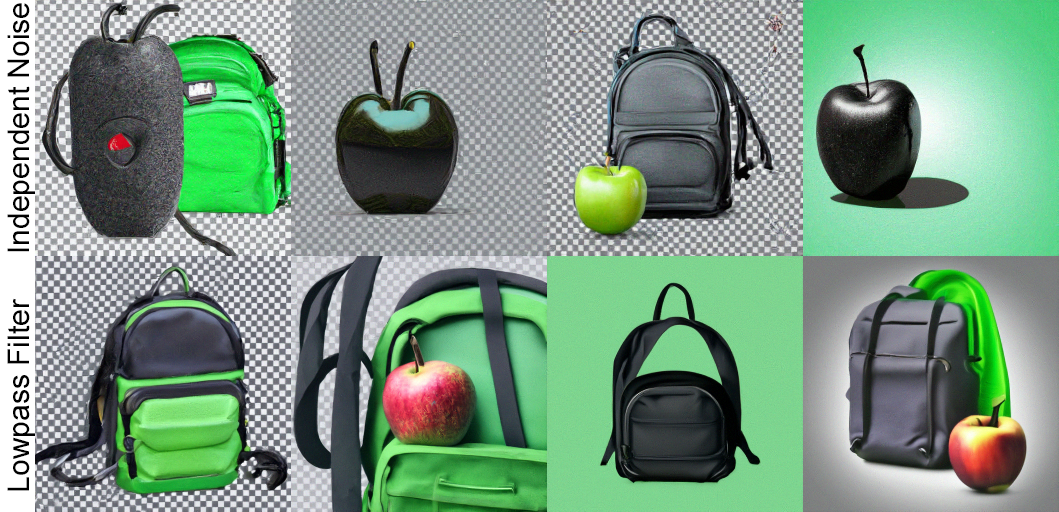}
\cprotect \caption{Outputs of layered $256 \times 256$ model from prompt \textit{A black apple and a green backpack.}}
\label{fig:noise_scale_comp}
\end{figure}
\begin{table}[H]
    \centering
    \begin{tabular}{llrl}
    \toprule
    \multicolumn{1}{c}{Noise Type} & \multicolumn{1}{c}{Target Resolution} &
    \multicolumn{1}{c}{FID} &
    \multicolumn{1}{c}{IS} \\
        \midrule
        Independent & $256 \times 256$ & $17.59$ & $29.32 \pm .46$ \\ 
        Sinc Interpolation & $256 \times 256$ & $\textbf{13.38}$ & $\textbf{29.62} \pm \textbf{.57}$ \\ 
        \midrule 
        Independent & $512 \times 512$ & $42.46$ & $28.89 \pm .22$ \\ 
        Sinc Interpolation & $512 \times 512$ & $\textbf{40.05}$ & $\textbf{28.74} \pm \textbf{.47}$ \\ 
        \bottomrule
    \end{tabular}
    \caption{\label{table:noise_scaling} Model performance for $256 \times 256$ and $512 \times 512$ layered models using independently sampled noise and scaled noise via sinc interpolation.}
\end{table}

\subsection{Cosine Schedule Shifting} \label{cosine_schedule_shifting_supp}

\begin{figure}[H]
\centering
\includegraphics[width=1\textwidth]{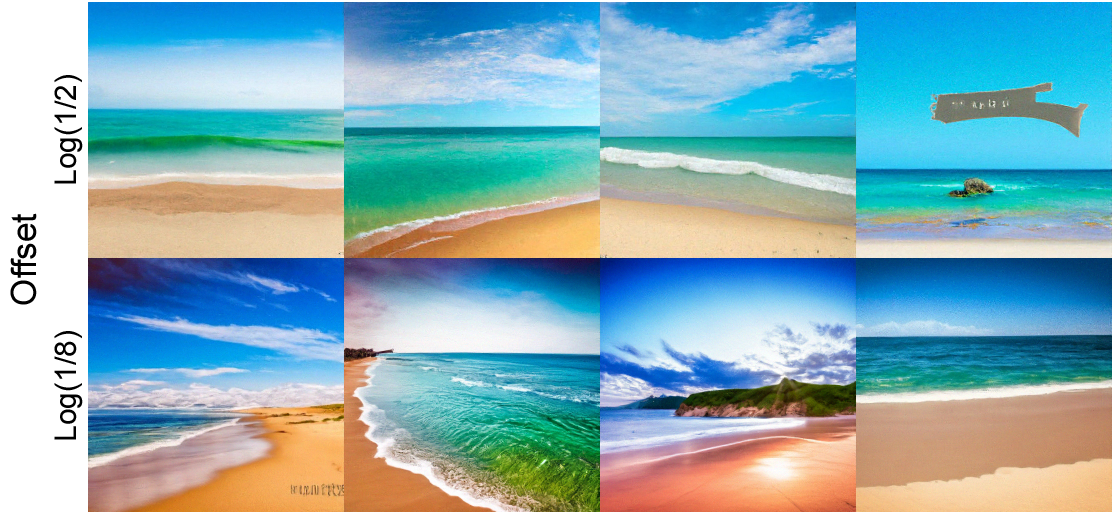}
\cprotect \caption{Outputs of layered $256 \times 256$ model from prompt \textit{landscape photo of beach with proof watermark}. We note that for a more aggressively shifted noise delay (bottom), higher resolution features can be seen in the waves. We also note the advent of a water-mark looking feature in the first image.}
\label{fig:256_noise_delay}
\end{figure}

\begin{figure}[H]
\centering
\includegraphics[width=1\textwidth]{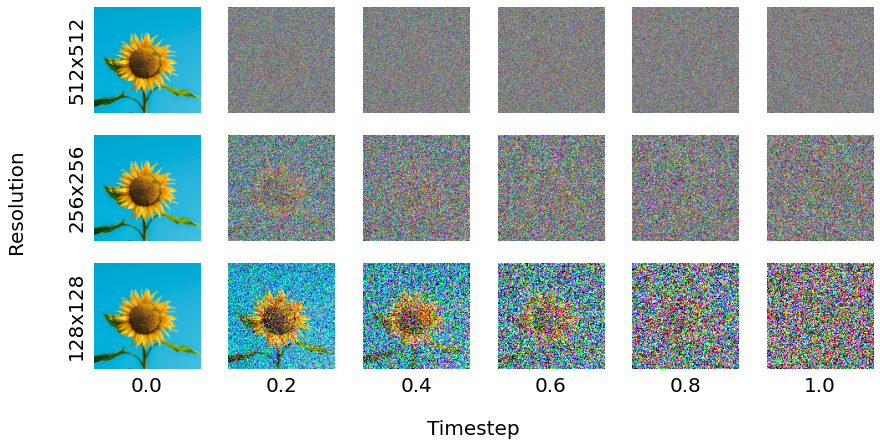}
\cprotect \caption{Demonstration of the increasingly shifted noise schedules for higher resolution on a reference image \cite{unsplash_photo_2020}. No shifting is applied to the cosine noise schedule at $128 \times 128$.}
\label{fig:cosine_schedule_shifting}
\end{figure}

\begin{figure}[H]
\centering
\includegraphics[width=1\textwidth]{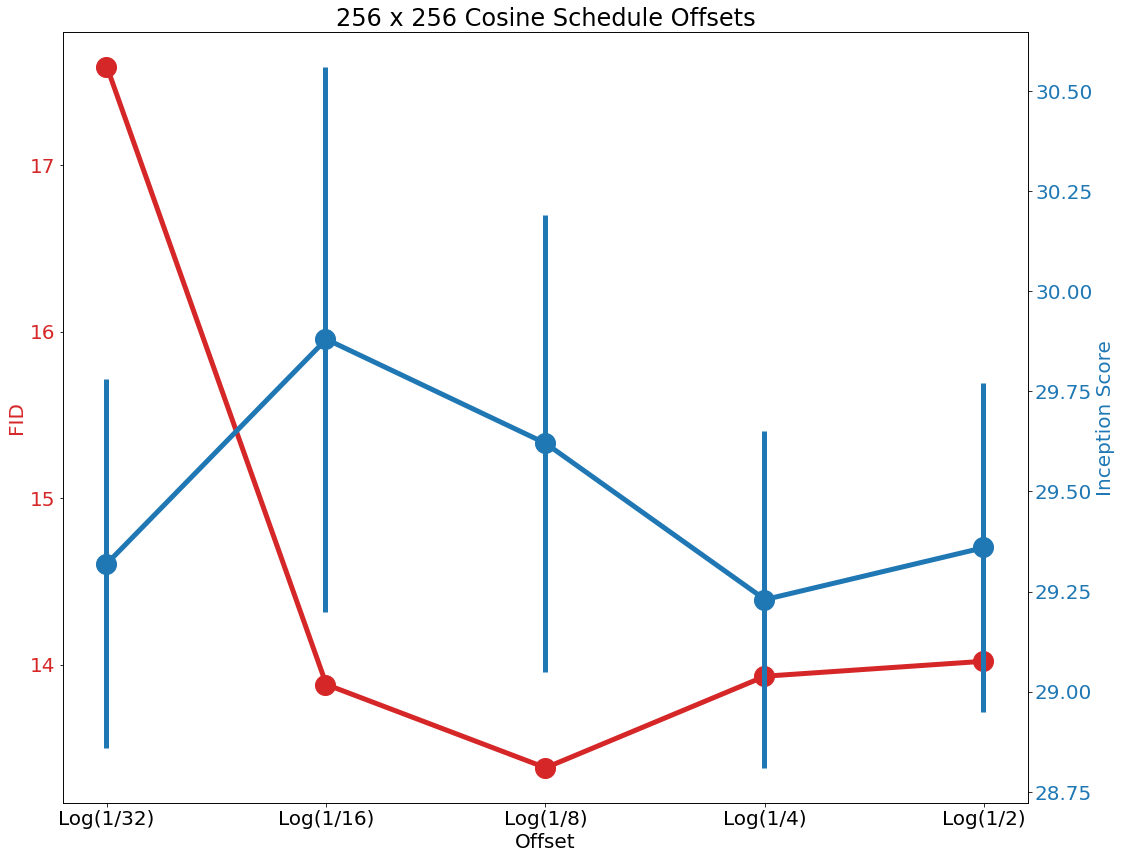}
\includegraphics[width=1\textwidth]{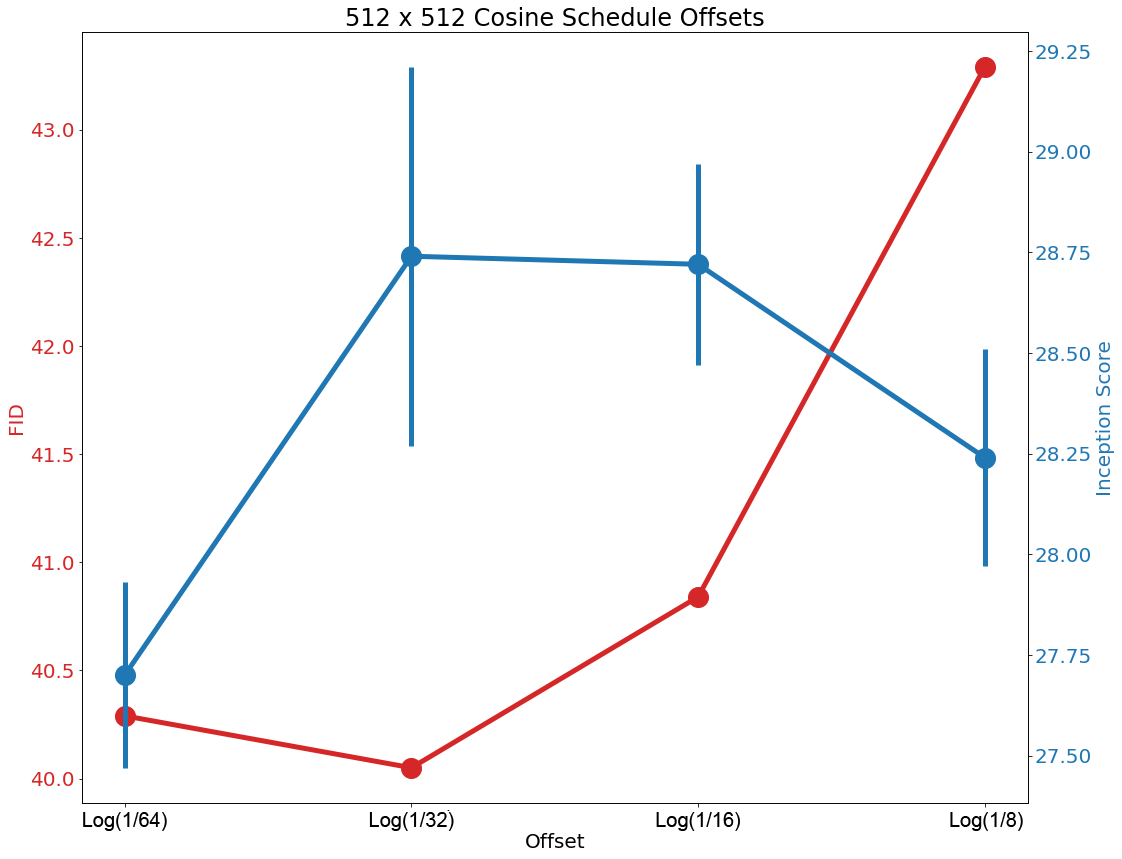}
\cprotect \caption{Sweeps showing IS and FID values on MSCOCO validation set for models trained with varying noise offsets. For the $512 \times 512$ model, we use a cosine schedule offset of $log\frac{1}{8}$ for the $256 \times 256$ layer.}
\label{fig:optimal_cosine_shift}
\end{figure}

\section{Training Optimizations} \label{training_optimizations_supp}

\subsection{Strategic Cropping} \label{cropping_supp}

\begin{figure}[H]
\centering
\includegraphics[width=1\textwidth]{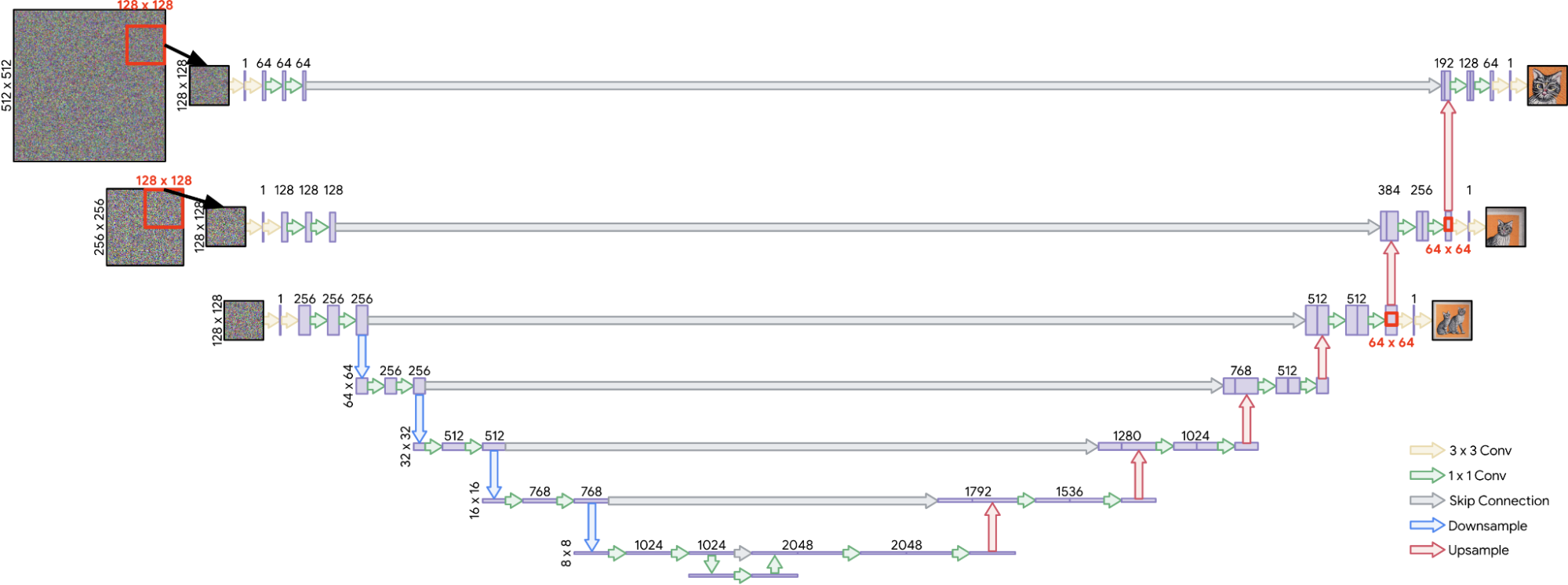}
\cprotect \caption{Model architecture with cropping applied. During training, all input images are $128 \times 128$. Observe in the upsampling stack, we take $64 \times 64$ crops prior to the upsampling convolution to ensure that the output is a $128 \times 128$ crop in the correct area.}
\label{fig:training_architecture}
\end{figure}

\subsection{Model Stacking} \label{model_stacking_supp}

\begin{table}[H]
    \centering
    \begin{tabular}{llrr}
    \toprule
    \multicolumn{1}{c}{Model} & \multicolumn{1}{c}{Resolution} &
    \multicolumn{1}{c}{FID} &
    \multicolumn{1}{c}{IS} \\
        \midrule
        (a) Random Initialization & $256 \times 256$ & $\textbf{13.38}$ & $\textbf{29.62} \pm \textbf{ .57}$ \\ 
        (b) Load $128 \times 128$ Model & $256 \times 256$ & $13.91$ & $28.85 \pm .03$ \\ 
        \midrule 
        Random Initialization & $512 \times 512$ & $\textbf{40.05}$ & $\textbf{28.74} \pm \textbf{ .47}$ \\ 
        Load (a) & $512 \times 512$ & $42.21$ & $27.82 \pm .21$ \\ 
        Load (b) & $512 \times 512$ & $43.53$ & $26.85 \pm .26$ \\ 
        \bottomrule
    \end{tabular}
    \caption{\label{table:model_stacking} Model performance for $256 \times 256$ and $512 \times 512$ layered models utilizing different amount of pre-training. Despite parts of the model essentially encountering a higher number of training steps and thereby images, we see degraded FID and IS.}
\end{table}

\begin{figure}[H]
\centering
\includegraphics[width=1\textwidth]{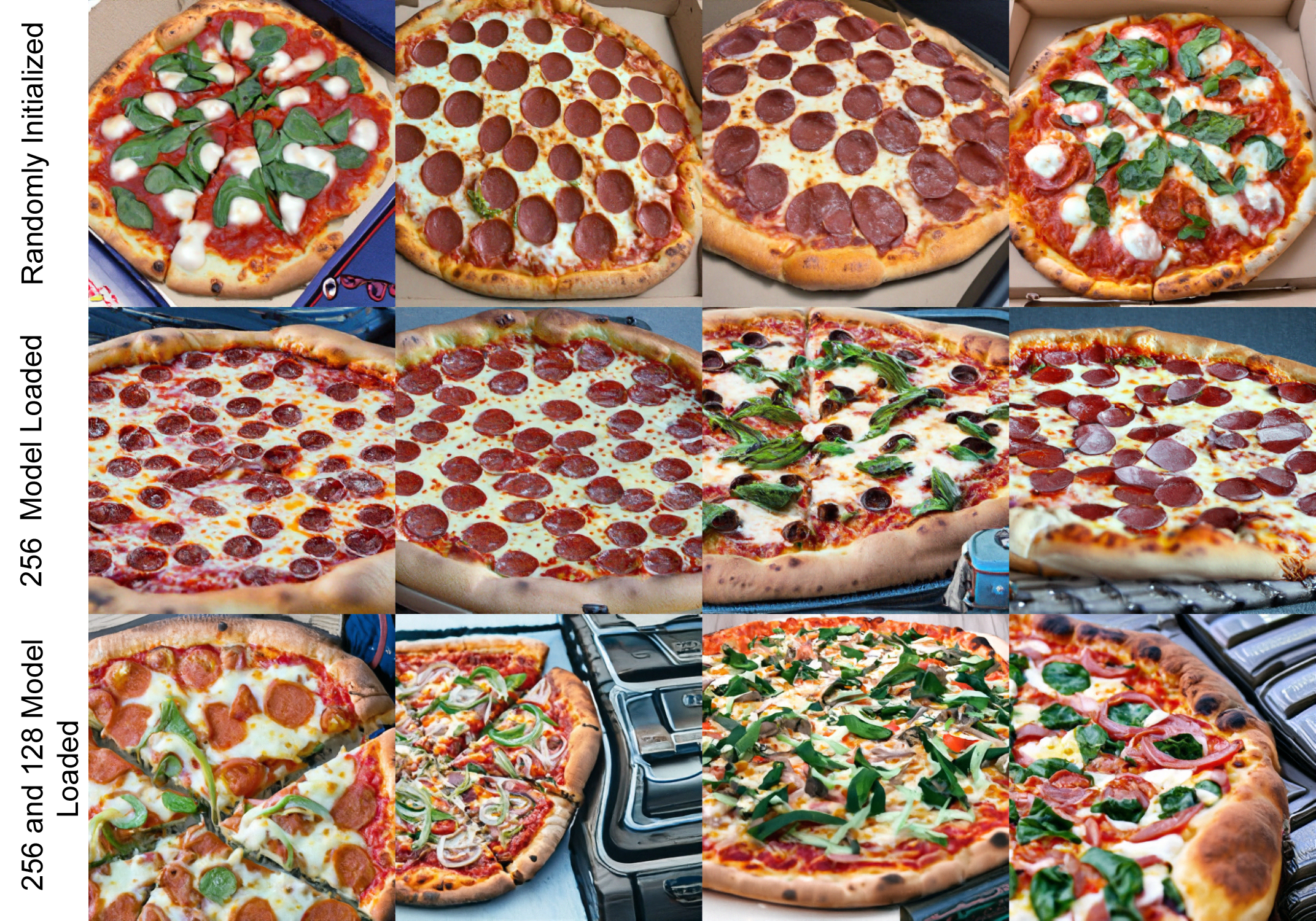}
\cprotect \caption{$512 \times 512$ output images for prompt \textit{A pizza on the right of a suitcase}. We note decreased image quality with increased initialization. However, we observe the opposite trend from the perspective of image alignment, where only the most initialized model generates the suitcase.}
\label{fig:pizza_comparison}
\end{figure}

{
    \small
    \bibliographystyle{cvpr/ieeenat_fullname}
    \bibliography{references}
}